\documentclass[lettersize,journal]{IEEEtran}
\usepackage{amsmath,amsfonts}
\usepackage{algorithmic}
\usepackage{array}
\usepackage[caption=false,font=normalsize,labelfont=sf,textfont=sf]{subfig}
\usepackage{textcomp}
\usepackage{stfloats}
\usepackage{url}
\usepackage{verbatim}
\usepackage{graphicx}
\usepackage{booktabs} 
\usepackage{booktabs}
\usepackage{makecell}
\usepackage{xcolor}

\newif\ifredmark   
\redmarkfalse
\newcommand{\rev}[1]{%
  \ifredmark
    \textcolor{red}{#1}%
  \else
    #1%
  \fi
}     
\usepackage{enumitem}

\def\BibTeX{{\rm B\kern-.05em{\sc i\kern-.025em b}\kern-.08em
    T\kern-.1667em\lower.7ex\hbox{E}\kern-.125emX}}
\usepackage{balance}
\begin{document}
\title{SMART: Shot-Aware Multimodal Video Moment Retrieval with Audio-Enhanced MLLM}

%


\author{An Yu, Weiheng Lu, Jian Li, Zhenfei Zhang, Yunhang Shen, 
Felix X.-F. Ye, Ming-Ching Chang
\thanks{An Yu is with the Department of Computer Science, University at Albany - SUNY, Albany, NY 12222, USA (e-mail: ayu@albany.edu).}
\thanks{Weiheng Lu is with the School of Software \& Microelectronics, Peking University, Haidian District, Beijing, China, 100871 (e-mail: luweiheng@stu.pku.edu.cn).}
\thanks{Jian Li is with Nanjing University, Nanjing, Jiangsu, China, 210093 (e-mail: swordlidev@gmail.com).}
\thanks{Zhenfei Zhang is with the Department of Computer Science, University at Albany - SUNY, Albany, NY 12222, USA (e-mail: zzhang45@albany.edu).}
\thanks{Yunhang Shen is with Xiamen University, Xiamen, Fujian, China, 361005 (e-mail: shenyunhang01@gmail.com).}
\thanks{ Felix X.-F. Ye is with the Department of Mathematics and Statistics, University at Albany - SUNY, Albany, NY 12222, USA (e-mail: xye2@albany.edu).}
\thanks{An Yu and Weiheng Lu contributed {\bf equally} to this work.}
\thanks{Ming-Ching Chang is the {\bf corresponding author}; he is with the Department of Computer Science, University at Albany - SUNY, Albany, NY 12222, USA (e-mail: mchang2@albany.edu).}

}


\maketitle

\begin{abstract}

Video Moment Retrieval aims to localize a target temporal segment in an untrimmed video given a natural-language query. Despite recent progress in both traditional approaches and multimodal large language model (MLLM)-based methods, many existing methods still rely primarily on visual cues and only weakly exploit temporal structure, which limits performance on complex videos. To address this issue, we introduce \textit{S}hot-aware \textit{M}ultimodal \textit{A}udio-enhanced \textit{R}etrieval of \textit{T}emporal \textit{S}egments (SMART), an MLLM-based framework that jointly leverages audio cues and shot-level temporal structure. SMART enriches multimodal representations by incorporating complementary audio information, and introduces a \textbf{Shot-aware Token Compression} strategy that preserves representative keyframes as full frames while compressing only non-keyframes to reduce visual redundancy. We also refine the prompting strategy to better utilize audio-visual context. Experiments on Charades-STA and QVHighlights show that SMART consistently improves over strong baselines and achieves favorable efficiency--effectiveness trade-offs, including gains of 1.61\% in R1@0.5 and 2.59\% in R1@0.7 on Charades-STA.

\end{abstract}


\section{Introduction}
\label{sec:intro}



\begin{figure}[t]
\centerline{
  \includegraphics[width=\columnwidth]{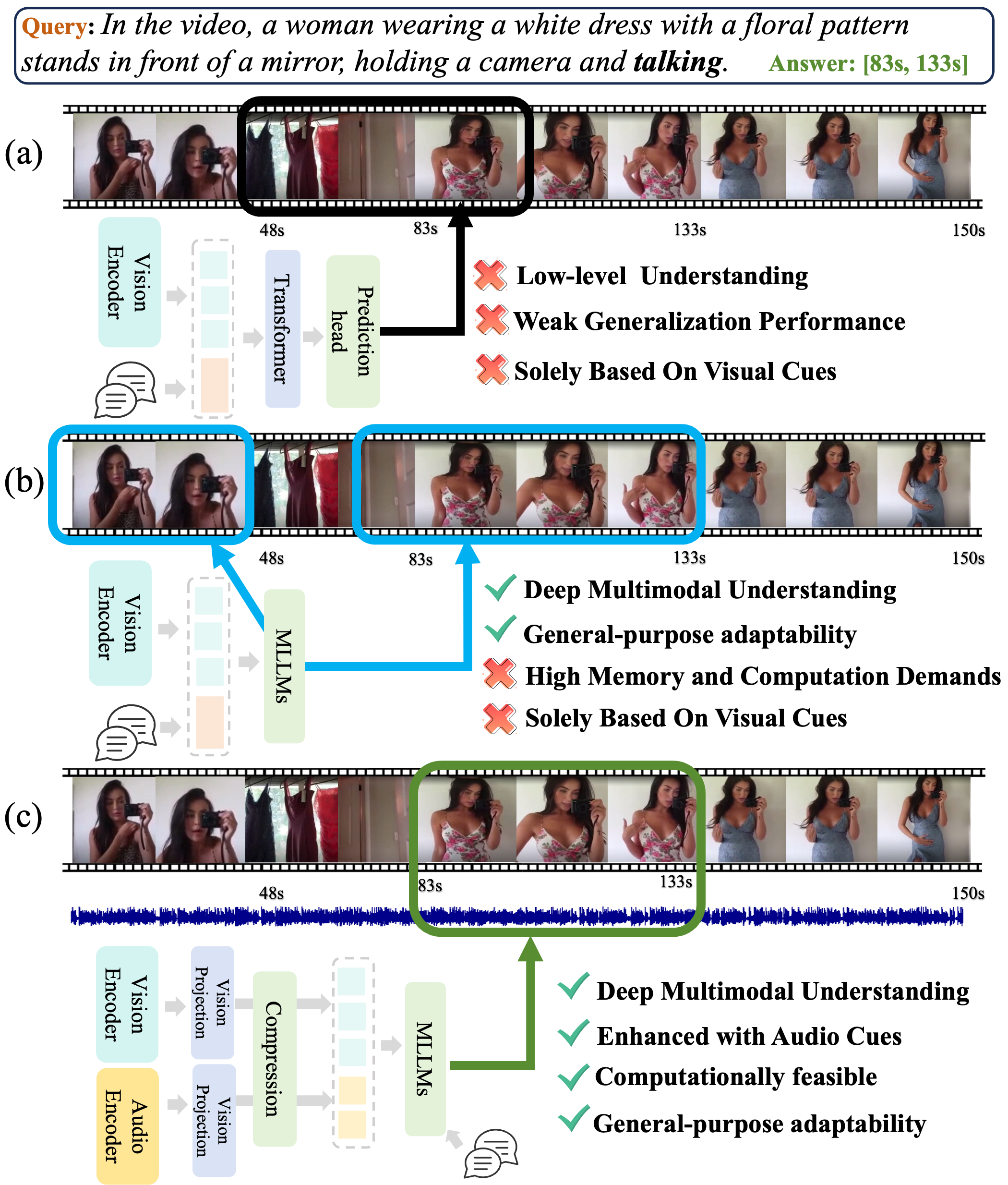} 
  \vspace{-2mm}
}
\caption{
{\bf Overview:} (a) Traditional models rely on low-level visual features and generalize poorly. (b) MLLM-based models improve semantic understanding but ignore audio and incur high computational cost. (c) Our {\bf SMART} model integrates audio and shot-coherent token compression for efficient, accurate multimodal moment retrieval.}
\label{fig:teaser} 
\end{figure}

\begin{figure*}[t] 
\centerline{
  \includegraphics[width=\linewidth]{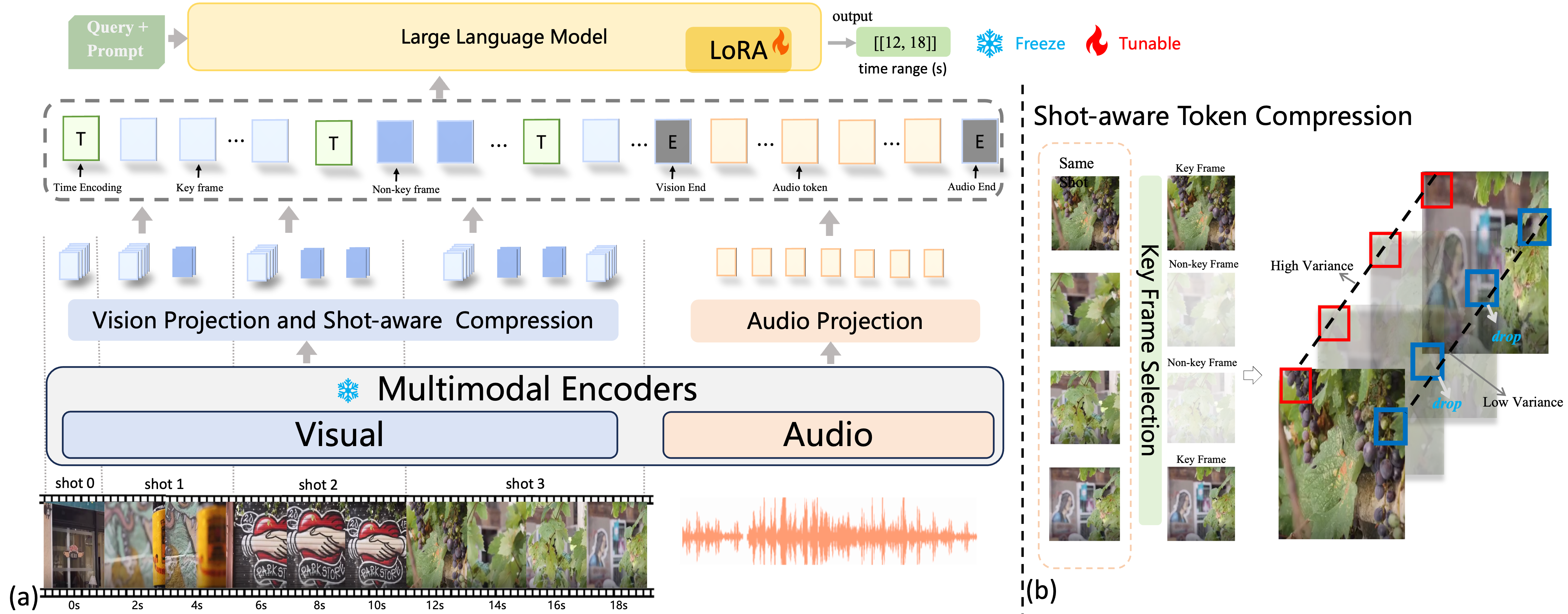}
  \vspace{-2mm}
}
\caption{
The {\bf Shot-aware Multimodal Audio-enhanced Retrieval of Temporal
Segments (SMART) architecture:} 
\textnormal{(a) The SMART pipeline integrates a pretrained MLLM with frozen visual and audio encoders and a lightweight LoRA-tuned LLM to predict temporal segments relevant to a query. Visual and audio features are projected, temporally encoded, and concatenated into a unified multimodal prompt, with shot-aware token compression applied for efficiency.
(b) The Shot-aware Token Compression (STC, detailed in Fig.~\ref{fig:compression}) operates in two stages: frames are first classified into key and non-key frames based on inter-frame differences; then, token-wise variance within each shot is analyzed. High-variance tokens (red), representing dynamic content, are retained, while low-variance tokens (blue) from non-key frames are discarded to remove redundancy while preserving essential cues.
}}
\label{fig:pipeline}
\end{figure*}

\IEEEPARstart{W}{ith} the rapid growth of video content shared and created on the internet and social media, the ability to efficiently analyze such content has become increasingly important. One key task in this domain is {\bf moment retrieval}---the process of identifying the specific temporal segment within a video that best corresponds to a given natural language query. This task is fundamental to a wide range of applications, including video summarization, content recommendation, security surveillance, and event localization. It has emerged as a prominent area of research due to its practical importance and inherent complexity.

Current moment retrieval methods can be broadly categorized into two main classes: traditional approaches~\cite{lei2021detecting, moon2024CGDETR, zeng2024UniMD},~\cite{snoek2007learned},~\cite{chen2023cross} and approaches that leverage Multimodal Large Language Models (MLLM)~\cite{jin2024efficientmllm, li2024surveybenchmarksmultimodallarge}. 
As shown in Figure~\ref{fig:teaser}\rev{(a)}, traditional approaches typically adopt Transformer-based architectures with task-specific prediction heads. These methods often rely on low-level visual features and are tailored to specific tasks, which limits their ability to generalize and capture high-level semantics. In contrast, recent studies~\cite{wang2024VideoAgent, liu2024LMR, lu2024llava, meinardus2024Mr.Blip, zhu2005insightvideo} have explored the use of MLLM to enhance semantic understanding for moment retrieval tasks. However, as illustrated in Figure~\ref{fig:teaser}\rev{(b)}, these methods typically rely solely on visual information and overlook the audio modality, which can provide complementary evidence and is particularly beneficial for queries involving sound or temporal continuity. For example, retrieving the moment for the query {\em ``a woman is holding a camera and talking''} requires not only visual identification but also speech detection. Similarly, the sound of a siren is often key to identifying moments involving passing vehicles. Without audio, models may miss complementary cues that are helpful for temporally ambiguous queries. This unimodal assumption poses challenges, especially when temporal grounding depends on acoustic cues or joint audio-visual signals.

LLMs often face challenges when handling long input sequences, which restricts the amount of video content they can effectively process and leads to high memory and computational demands. As a result, they may fail to capture the complete and nuanced temporal context spread across extended video sequences, which is essential for accurately identifying and retrieving the segments that align closely with a given natural language query. The inefficiency in handling long sequences also poses a major scalability challenge, especially in real-world applications where videos are lengthy and complex. These constraints reduce their capacity to capture rich, multimodal temporal context, thus limiting their effectiveness in tasks that require precise and robust video understanding and moment retrieval.

To address limitations in multimodal modeling and efficiency for long-video moment retrieval, we propose \textbf{S}hot-aware \textbf{M}ultimodal \textbf{A}udio-enhanced \textbf{R}etrieval of \textbf{T}emporal Segments (SMART)—a novel MLLM-based framework illustrated in Figure~\ref{fig:teaser}\rev{(c)}). SMART integrates audio information and employs a dual-branch encoder to jointly process video and audio inputs, enriching multimodal representations and enabling more comprehensive temporal understanding. To reduce the computational cost of long input sequences, we introduce {\bf Shot-aware Token Compression (STC)} that compresses redundant frame tokens within each shot. By computing token-level variance across frames using Q-Former outputs, it identifies and retains key tokens while discarding redundant ones. This compression aims to preserve semantic richness and fine-grained temporal cues, allowing SMART to process longer video sequences more efficiently while reducing redundant visual tokens. The complete pipeline is shown in Figure~\ref{fig:pipeline}.

We conduct extensive experiments on two widely used benchmark datasets, Charades-STA~\cite{gao2017tall} and QVHighlights~\cite{lei2021detecting}. The results show that SMART consistently improves over strong baselines by better leveraging complementary audio cues and reducing visual redundancy through shot-aware compression. On the QVHighlights test set, SMART improves R1@0.5 by 1.56\% over LLaVA-MR. On Charades-STA, it improves R1@0.5 by 1.61 points and R1@0.7 by 2.59\% over LLaVA-MR.

The main contributions of this paper are summarized as follows:
\begin{itemize}
    \item We propose SMART, a multimodal large language model (MLLM)-based framework for moment retrieval that incorporates complementary audio cues to support temporal localization.

    \item We introduce a shot-aware token compression strategy that preserves representative keyframes as full frames and compresses only non-keyframes, thereby reducing visual redundancy while retaining informative temporal evidence.

    \item We conduct comprehensive experiments on Charades-STA and QVHighlights, showing that SMART achieves consistent performance gains and favorable efficiency--effectiveness trade-offs over strong baselines.
\end{itemize}


\rev{\section{Related Work}}
\subsection{Video Moment Retrieval}

Video Moment Retrieval (VMR), also known as temporal sentence grounding in videos, aims to localize the video segment that best matches a natural-language query. Early methods such as TALL~\cite{gao2017tall} formulated this task as sliding-window matching, while attention-based designs improved efficiency by directly predicting temporal boundaries~\cite{anne2017localizing}. Subsequent approaches strengthened video--text alignment through query-conditioned modulation and structured reasoning, as represented by Semantic Conditioned Dynamic Modulation~\cite{yuan2019semantic} and the Moment Alignment Network~\cite{zhang2019man}. More recent studies have explored stronger temporal modeling, supervision-efficient localization, and generalization-oriented settings, including frame-supervised localization~\cite{liu2025gaming} and zero-shot VMR~\cite{jiang2024zero}. Despite these advances, most VMR methods still focus primarily on visual-text matching and treat videos as frame sequences or clip sequences without explicitly modeling shot-level organization. This can introduce substantial visual redundancy, especially for long videos with repeated or visually similar frames. SMART addresses this gap by integrating audio cues and shot-aware compression within an MLLM-based retrieval framework, improving temporal grounding while reducing unnecessary visual tokens.

\subsection{Audio-Enhanced Video Understanding and Retrieval}

Audio provides complementary temporal and semantic evidence for video understanding, especially when the target event is only partially visible or when key cues are conveyed by sound. Recent audio-aware VMR work~\cite{lin2025audio} shows that acoustic information can materially improve localization for sound-related queries. Beyond VMR, audio-visual event perception and parsing methods further demonstrate the value of acoustic cues in noisy or weakly supervised settings~\cite{jiang2025resisting, gao2025learning}. However, audio is not uniformly beneficial: its utility depends on whether the query is acoustically grounded, whether the sound is temporally aligned with the visual event, and how audio is fused with visual and textual information. Different from prior audio-enhanced methods, SMART studies audio within an MLLM-based retrieval pipeline and combines it with shot-aware visual compression. This design allows the model to retain richer multimodal context while limiting redundancy in long video sequences.

\subsection{Robustness and Debiasing in Temporal Grounding}

Robustness and debiasing are also closely related to temporal grounding. Existing datasets and models may contain temporal shortcuts, annotation artifacts, or evaluation biases, causing models to exploit spurious correlations rather than genuine temporal reasoning. Prior studies have therefore examined dataset bias, evaluation bias, counterfactual analysis, and debiasing strategies for temporal sentence grounding~\cite{lan2023closer, qi2024bias, jiang2024counterfactually, kong2025reverse}. These works show that strong in-distribution performance does not necessarily indicate robust temporal understanding.

Although SMART does not directly target dataset debiasing, its design is related to robustness in two ways. First, shot-aware compression reduces dense and redundant visual evidence, encouraging the model to focus on temporally informative content instead of overusing frame-level repetition. Second, our analyses examine sensitivity to shot-boundary perturbations and coarse audio temporal shifts, which provides insight into how multimodal temporal grounding behaves under structured perturbations and modality-specific noise. This is important because small changes in temporal segmentation or modality alignment may alter the evidence available to the model without changing the underlying query intent. \rev{Recent robust and data-efficient adaptation studies further show that noisy samples, channel bias, and feature redundancy can reduce the reliability of pretrained visual representations~\cite{zhang2025reliable, zhang2026channel}. Although these works mainly study few-shot visual recognition rather than VMR, they support our motivation to reduce noisy or redundant evidence through structured compression.}

\subsection{Multimodal LLMs for Video Understanding}

Building on the success of large language models, Multimodal Large Language Models (MLLMs) have become a powerful paradigm for cross-modal reasoning. Existing MLLMs extend LLMs with gated cross-attention, adapter modules, projection layers, or Q-Formers to bridge visual and textual representations~\cite{alayrac2022flamingo, awadalla2023OpenFlamingo, li2023MIMIC-IT, zhang2024LLaMA-Adapter, dai2023InstructBLIP, liu2023LLaVA, zhu2023MiniGPT-4}. Video-oriented MLLMs further adapt these mechanisms to temporal inputs and support video-level reasoning~\cite{li2024VideoChat, zhang2023Video-LLaMA, maaz2024Video-ChatGPT, luo2023Valley, ji2024wavchat}. Recent retrieval-oriented models such as Mr.~BLIP~\cite{meinardus2024Mr.Blip} and LLaVA-MR~\cite{lu2024llava} demonstrate that MLLMs can improve semantic alignment for moment retrieval. \rev{In parallel, prompt-tuning methods study parameter-efficient adaptation and base-to-new generalization for vision-language models~\cite{zhang2026conditional}, which is complementary to our focus on audio-enhanced VMR with shot-aware compression.} Nevertheless, directly applying MLLMs to long videos remains challenging because dense visual tokens increase computation and may dilute temporally relevant evidence.

Our design is also related to ActBERT~\cite{zhu2020actbert}, which learns joint video-text representations by combining global action information with local object-level regions. Although ActBERT predates recent MLLMs and does not address audio-enhanced moment retrieval, its global-local design philosophy is aligned with SMART's goal of preserving coarse temporal organization while reducing local redundancy. Unlike prior MLLM-based retrieval methods that mainly emphasize semantic alignment, SMART further incorporates explicit audio integration and shot-aware visual compression. As a result, it seeks a better balance between multimodal reasoning, temporal structure preservation, and computational efficiency for video moment retrieval.



\section{Method}

We advance MLLMs for moment retrieval by incorporating audio as complementary contextual information alongside visual inputs, together with a token compression mechanism to reduce sequence length. The following sections present the overall SMART architecture ($\S$~\ref{sec:process_pipeline}), describe the audio integration strategy that enhances multimodal understanding ($\S$~\ref{sec:audio_intergration}), and introduce a shot-aware token compression approach designed to reduce computational cost while preserving fine-grained temporal detail ($\S$~\ref{sec:shot_tc}).

\subsection{Pipeline Design}
\label{sec:process_pipeline}

Our model architecture is illustrated in Fig.~\ref{fig:pipeline}. The model uses two separate branches for visual and audio feature extraction. In the visual branch, EVA-CLIP~\cite{sun2023evaclipimprovedtrainingtechniques} is adopted as the visual encoder to process sampled frames and produce a sequence of image embeddings, denoted as $\mathbf{V} = [\mathbf{v}_1, \dots, \mathbf{v}_N] \in \mathbf{R}^{N \times P \times D_v}$, where $N$ is the number of frames, $P$ is the number of patches per frame, and $D_v$ is the visual feature dimension. A query-based transformer (Q-Former) then refines each frame embedding to generate frame-level features $\mathbf{V}^q = [\mathbf{v}_1^q, \mathbf{v}_2^q, \dots, \mathbf{v}_N^q] \in \mathbf{R}^{N \times Q \times D_q}$, with $Q$ queries and a Q-Former output dimension of $D_q$. The output, $\mathbf{V}^q$, is projected into the LLM feature space via a vision projection layer. The shot-coherent token compression module then reduces token redundancy, producing the final visual representation $\mathbf{V}^L \in \mathbf{R}^{S_v \times D_L}$, where $S_v$ is the number of compressed visual tokens and $D_L$ is the LLM input dimension.

In the audio branch, we adopt BEATs~\cite{chen2022BEATs}, an acoustic tokenizer that produces discrete embeddings rich in semantic information. The audio embeddings are denoted as $\mathbf{A} = [\mathbf{a}_1, \dots, \mathbf{a}_T] \in \mathbf{R}^{T \times D_a}$, where $T$ is the number of audio tokens and $D_a$ is the audio feature dimension. Given the continuous but often redundant nature of audio data, we apply average pooling to reduce the sequence length, followed by an audio projection layer that maps the audio features into the LLM space. The final audio representation is $\mathbf{A}^L \in \mathbf{R}^{S_a \times D_L}$, where $S_a$ is the pooled length of audio tokens.

To construct the input for the language model, we organize a multimodal sequence consisting of time tokens $t_i$, visual tokens $\mathbf{v}_i^L \in \mathbf{V}^L$, audio tokens $\mathbf{a}_i^L \in \mathbf{A}^L$, a task-specific query $q$, and a prompt $p$. To provide explicit temporal context, each time token is placed directly before its corresponding frame embedding. We assign explicit temporal tokens only to visual frame embeddings, while audio features are incorporated as pooled acoustic context rather than being explicitly aligned with individual visual timestamps. Additionally, to distinguish between modalities, we insert special separator tokens: $V_E$ to mark the end of visual embeddings and $A_E$ for the end of audio embeddings. The resulting multimodal input sequence is structured as:
\begin{equation}
x = [t_1, \mathbf{v}_1^L, t_2, \mathbf{v}_2^L, \dots, V_E, \mathbf{a}_1^L, \mathbf{a}_2^L, \dots, A_E, q, p]
\end{equation}
This token sequence is then fed into the LLM to identify time segments relevant to the input query $q$. Inspired by the design of Mr. BLIP~\cite{meinardus2024Mr.Blip}, the model is prompted to generate a sequence of one or more temporal moments, each represented by a start and end timestamp in seconds. The output follows a nested list format, {\em e.g.}, \texttt{[[s$_1$, e$_1$], [s$_2$, e$_2$], ...]}.

\subsection{Audio Integration}
\label{sec:audio_intergration}

Current mainstream benchmark datasets for moment retrieval predominantly provide textual descriptions grounded in visual content, leading most existing methods to rely primarily on visual cues. However, audio can offer complementary information to the visual scene; for instance, footsteps may provide useful acoustic evidence for walking-related actions. Incorporating such acoustic cues enables richer multimodal reasoning and can improve temporal localization, particularly in cases where visual evidence alone is insufficient or temporally ambiguous.

To incorporate audio, we introduce an efficient module that extracts semantic acoustic cues and injects them as complementary context into the vision-language framework. Raw audio is extracted and resampled to a standardized 16kHz rate, then encoded using the pretrained BEATs~\cite{chen2022BEATs} model, which is known for its strong semantic understanding of audio signals. To reduce redundancy and maintain computational efficiency, the audio features are temporally pooled into a compact representation and projected to match the visual feature dimension. This design facilitates seamless cross-modal integration, allowing the model to leverage auditory cues (such as speech or environmental sounds) for improved temporal grounding.

Notably, we do not explicitly insert timestamp tokens into the audio features. Instead, audio is incorporated as complementary contextual information rather than an explicitly time-aligned stream. BEATs provides temporally ordered acoustic representations through frame-wise encoding and positional embeddings, which are sufficient for our design without requiring explicit fine-grained audio–visual alignment. We further explore and compare different audio integration strategies in the Ablation Study in $\S$~\ref{sec:ablation}.

\begin{figure}[t]
\centerline{
  \includegraphics[width=\linewidth]{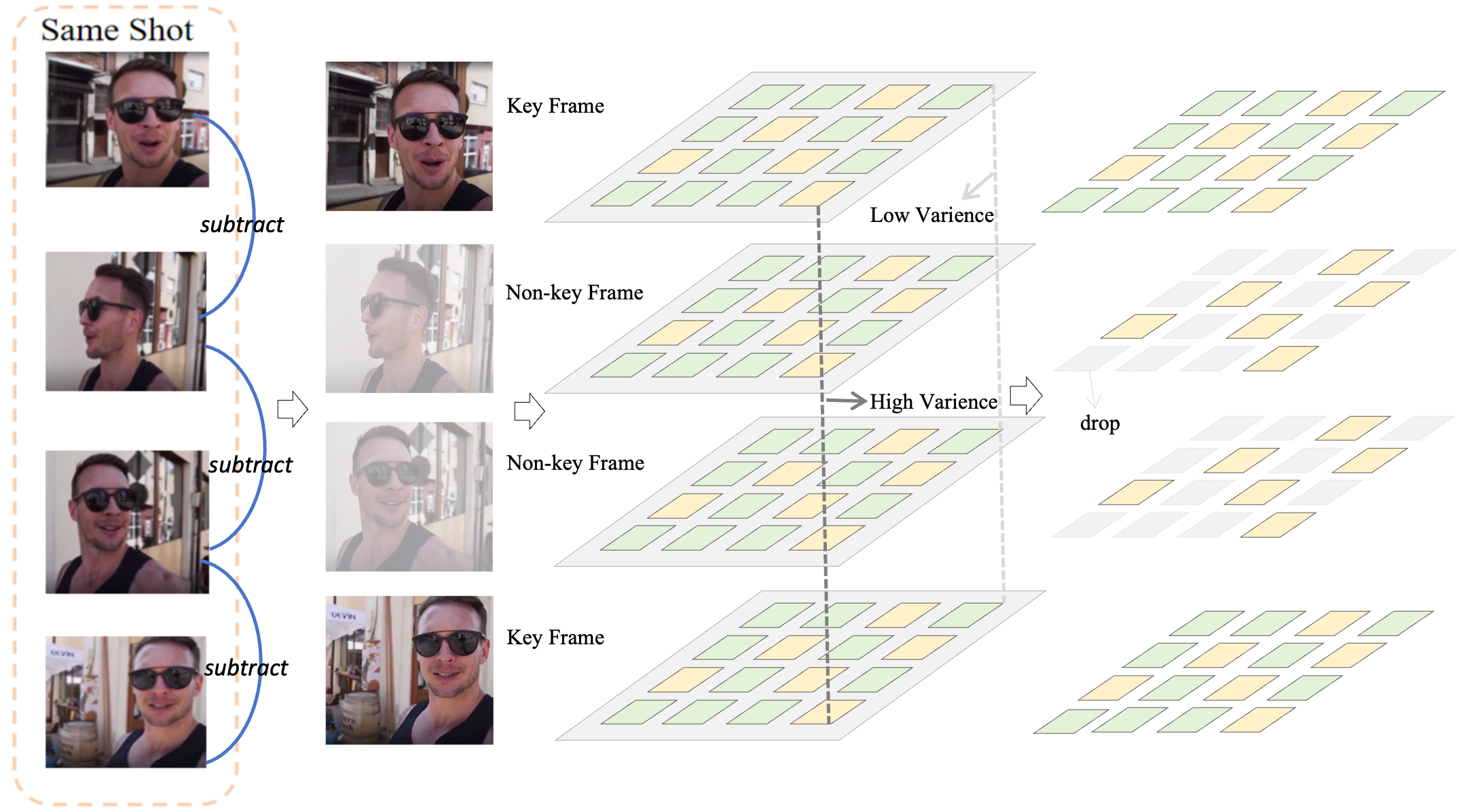}
  \vspace{-2mm}  
}
\caption{
{\bf Shot-aware token compression:}
{\normalfont Key frames are detected via inter-frame differences. Within each shot, high-variance tokens from dynamic regions are preserved, while low-variance tokens from non-key frames are discarded, effectively minimizing redundancy and retaining essential temporal cues for accurate moment retrieval.
}}
\label{fig:compression} 
\vspace{-5mm}
\end{figure}

\subsection{Shot-aware Token Compression (STC)}
\label{sec:shot_tc}

We propose the {\bf Shot-aware Token Compression (STC)} strategy to efficiently process long multimodal videos while preserving fine-grained temporal cues. The key insight is that frames within the same shot exhibit strong visual similarity and temporal coherence, introducing redundancy that can be exploited for compression. STC addresses this by (1) identifying keyframes that capture meaningful scene dynamics, and (2) performing intra-shot token compression that removes redundant visual tokens from non-keyframes. This design substantially reduces sequence length while maintaining essential spatio-temporal semantics for accurate moment retrieval. For guiding the compression process, we first segment each video into distinct shots using TransNetV2~\cite{souček2020transnetv2effectivedeep}, which detects shot boundaries based on learned visual transition cues.

\medskip
\noindent
\textbf{Motivation:} Dense frame sampling is crucial for capturing subtle visual transitions and transient events, yet it greatly increases sequence length—particularly when combined with audio inputs. Transformer-based LLMs suffer from quadratic self-attention complexity, resulting in high memory cost and degraded inference efficiency for long sequences. Thus, effective token compression is necessary to remove redundancy while preserving discriminative temporal and semantic information.
Existing approaches, such as clustering or pooling~\cite{zhao2022centerclip,clavie2024reducing}, often lose fine details in dynamic scenes. In contrast, STC performs compression at the shot level, where intra-shot visual redundancy is high and temporal coherence is preserved, ensuring better efficiency–accuracy trade-offs.

\medskip
\noindent
\textbf{Stage 1 Keyframe Identification:}
Given visual features $\mathbf{F}^v$ extracted from the image encoder, we compute frame-to-frame differences to detect visually dynamic frames: $\Delta \mathbf{V} = [\Delta \mathbf{v}_1, \dots, \Delta \mathbf{v}_N]$, where each $\Delta \mathbf{v}_i$ encodes patch-wise changes between consecutive frames.
The L2 norm of each $\Delta \mathbf{v}_i$ yields a motion magnitude sequence $\mathbf{d} = [d_1, d_2, \dots, d_N]$. To suppress local jitter and noise, we apply a Gaussian filter to obtain a smoothed series $\hat{\mathbf{d}} = \text{Gaussian}(\mathbf{d})$, where each $\hat{d}_i$ represents the smoothed motion magnitude of the $i$-th frame.
Frames with higher $\hat{d}_i$ values are selected as keyframes that contain dynamic content, while the remaining frames are considered non-keyframes with redundant information suitable for compression.

\medskip
\noindent
\textbf{Stage 2 Intra-Shot Token Compression:}
For each shot, token-level compression is performed within the Q-Former structure, which uses $Q$ learnable queries to encode each frame. Each query captures distinct semantic aspects of the visual content. For all frames in a shot, token variance across the temporal dimension is computed at each query position. High-variance tokens correspond to dynamic elements and are retained across all frames. Low-variance tokens, typically representing static background or redundant content, are discarded from non-keyframes. Their information can be inferred from corresponding tokens in keyframes.
This selective compression reduces visual redundancy and computational load, while preserving the temporal fidelity necessary for precise multimodal moment retrieval. Figure~\ref{fig:compression} illustrates the STC process.


\section{Experimental Evaluation}

\subsection{Experiment Setting}
\label{sec:dataset and setup}

\medskip
\noindent
\textbf{Benchmarks:} We validate our model on two widely-used video Moment Retrieval (MR) datasets: (1) {\bf Charades-STA}~\cite{gao2017tall} contains 9,848 videos with an average duration of 30.6 seconds. It provides 16,128 annotations (12,408 for training and 3,720 for testing), where moments average 8.1 seconds in length and queries average 7.22 words. (2) {\bf QVHighlights}~\cite{lei2021detecting} consists of 10,148 YouTube videos, each 150 seconds long. Every video includes at least one annotated query (averaging 11.3 words) describing relevant moments, which average 24.6 seconds in duration. The dataset is split into 7,218 training queries and 1,150 validation queries.

Although ActivityNet Captions~\cite{krishna2017Anet} is widely used for moment retrieval, we do not include it in our experiments because the commonly used official release (``Anet\_videos\_15fps\_short256'') contains silent videos only, which prevents us from evaluating the audio branch of SMART in its intended multimodal setting. Similarly, datasets such as TACoS~\cite{rohrbach2014coherent} and MAD~\cite{soldan2022mad} are typically provided as pre-extracted visual features without access to raw video frames or audio streams. Since SMART relies on frame-level visual processing, shot detection, and audio encoding as part of the full pipeline, these datasets do not support a fair and reproducible evaluation of the complete method. We therefore focus our experiments on benchmarks that allow the full multimodal pipeline to be executed. In particular, QVHighlights serves as our main long-video benchmark for analyzing the proposed audio design and shot-aware token compression.

\medskip
\noindent\textbf{Evaluation Metrics:}
We evaluate Moment Retrieval (MR) using standard metrics: Recall@K, mean Average Precision (mAP), and mean Intersection over Union (mIoU), computed at specific IoU thresholds. Recall@K measures the proportion of top-K predictions with IoU above a threshold (e.g., R1@0.5). mIoU calculates the average IoU across all predictions, while mAP measures the mean precision across multiple IoU thresholds (e.g., mAP@0.5, mAP@0.75). In summary, R1@0.5 and R1@0.7 assess retrieval accuracy at different IoU levels, mIoU captures average segment overlap, and mAP reflects overall localization precision across thresholds.

\medskip
\noindent\textbf{Implementation Details:} We adopt parameter-efficient fine-tuning with LoRA~\cite{hu2021lora}, updating only 0.63\% of the model parameters. To mitigate occasional inconsistencies in LLM outputs, we apply lightweight post-processing heuristics to correct minor formatting errors and improve prediction reliability. Training is performed with the AdamW optimizer~\cite{loshchilov2019AdamW}, starting from a learning rate (LR) of 1e-8, linearly warmed up to 3e-4 over the first 10\% of iterations, and subsequently decayed with a cosine schedule. During training, frames are randomly sampled from each video.

For Charades-STA, we sample 60 frames per video and train for 30 epochs with a batch size of 16 using 4 GPUs. For QVHighlights, we sample 80 frames per video and train for 30 epochs with a batch size of 32 on 8 GPUs, applying gradient accumulation with a factor of 4. All experiments are conducted on NVIDIA A100 GPUs with 80GB memory.

\begin{table}[t]
\caption{
Comparison with state-of-the-art methods on the Charades-STA and QVHighlights datasets. \textbf{CLIP*}: UnLoc~\cite{yan2023UnLoc-L} pretrained on Kinetics 400/700 action classification, \textbf{SF}: Slow-Fast backbone. 
\vspace{-2mm}
}
\label{tab:sota}
\centerline{
\setlength{\tabcolsep}{0.1mm}
\resizebox{1.0\linewidth}{!}{
\footnotesize
\begin{tabular}{lcccccc}
\toprule
Method & Backbone & mIoU & R1@0.5 & R1@0.7 & mAP@0.5 & mAP@0.75 \\
\midrule
\multicolumn{7}{c}{\textbf{QVHighlights Validation set}} \\
Moment-DETR~\cite{lei2021detecting} & SF+CLIP & -- & 59.68 & 40.84 & -- & -- \\
EaTR~\cite{jang2023EaTR} & I3D & -- & 61.36 & 45.79 & 61.86 & 41.91 \\
QD-DETR~\cite{moon2023QD-DETR} & SF+CLIP & -- & 62.68 & 46.66 & 62.23 & 41.82 \\
UVCOM~\cite{xiao2023UVCOM} & SF+CLIP & -- & 65.10 & 51.81 & -- & -- \\
UnLoc-L~\cite{yan2023UnLoc-L} & CLIP* & -- & 66.10 & 46.70 & -- & -- \\
CG-DETR~\cite{moon2024CGDETR} & SF+CLIP & -- & 67.40 & 52.10 & 65.60 & 45.70 \\
$R^2$-Tuning~\cite{liu2024r2$R^2$-Tuning} & CLIP & -- & 68.71 & 52.06 & -- & -- \\
VideoLights-B-pt~\cite{paul2024videolights} & SF+CLIP+BLIP & -- & 72.06 & 57.94 & 70.38 & 51.12 \\
FlashVTG~\cite{cao2024FlashVTG} & SF+CLIP & -- & 73.10 & 57.29 & 72.75 & 54.33 \\
Mr.BLIP~\cite{meinardus2024Mr.Blip} & BLIP-2 & -- & 76.13 & 63.35 & 69.39 & 55.78 \\
LLaVA-MR~\cite{lu2024llava} & CLIP+LLM & 71.85 & \underline{78.13} & \underline{64.13} & \underline{69.64} & \underline{56.32} \\
\midrule
\textbf{SMART} & CLIP+LLM & \textbf{72.03} & \textbf{78.65} & \textbf{65.03} & \textbf{70.46} & \textbf{56.72}\\
\midrule
\multicolumn{7}{c}{\textbf{QVHighlights Test Set}} \\
SeViLa~\cite{yu2023self} & BLIP-2 & -- & 54.50 & 36.50 & -- & -- \\
UniVTG~\cite{lin2023UniVTG} & SF+CLIP & -- & 58.86 & 40.86 & 57.60 & 35.59 \\
QD-DETR~\cite{moon2023QD-DETR} & SF+CLIP & -- & 62.40 & 44.98 & 62.52 & 39.88 \\
UVCOM~\cite{xiao2023UVCOM} & SF+CLIP & -- & 63.55 & 47.47 & 63.37 & 42.67 \\
LMR~\cite{liu2024LMR} & SF+CLIP & -- & 64.40 & 47.21 & 64.65 & 43.16 \\
CG-DETR~\cite{moon2024CGDETR} & SF+CLIP & -- & 65.43 & 48.38 & 64.51 & 42.77 \\
$R^2$-Tuning~\cite{liu2024r2$R^2$-Tuning} & CLIP & -- & 68.03 & 49.35 & 69.04 & 47.56 \\
VideoLights-B-pt~\cite{paul2024videolights} & SF+CLIP+BLIP & -- & 70.36 & 55.25 & 69.53 & 49.17 \\
FlashVTG~\cite{cao2024FlashVTG} & SF+CLIP & -- & 70.69 & 53.96 & 72.33 & 53.85 \\
Mr.BLIP~\cite{meinardus2024Mr.Blip} & BLIP-2 & -- & 74.77 & 60.51 & 68.12 & 53.38 \\
LLaVA-MR~\cite{lu2024llava} & CLIP+LLM & -- & \underline{76.59} & \underline{61.48} & \underline{69.41} & \underline{54.40} \\
\midrule
\textbf{SMART} & CLIP+LLM & -- & \textbf{78.15} & \textbf{63.16} & \textbf{70.76} & \textbf{55.54} \\
\midrule
\multicolumn{7}{c}
{\textbf{Charades-STA Test set}} \\
Moment-DETR~\cite{lei2021detecting} & & -- & 53.63 & 31.37 & -- & -- \\
QD-DETR~\cite{moon2023QD-DETR} & SF+CLIP & -- & 57.31 & 32.55 & -- & -- \\
LMR~\cite{liu2024LMR} & SF+CLIP & -- & 55.91 & 35.19 & -- & -- \\
UniVTG~\cite{lin2023UniVTG} & SF+CLIP & 50.10 & 58.01 & 35.65 & -- & -- \\
CG-DETR~\cite{moon2024CGDETR} & SF+CLIP & 50.13 & 58.44 & 36.34 & -- & -- \\
UVCOM~\cite{xiao2023UVCOM} & SF+CLIP & -- & 59.25 & 36.64 & -- & -- \\
$R^2$-Tuning~\cite{liu2024r2$R^2$-Tuning} & CLIP & -- & 59.78 & 37.02 & -- & -- \\
UnLoc-L~\cite{yan2023UnLoc-L} & CLIP* & -- & 60.80 & 38.40 & -- & -- \\
VideoLights-B-pt~\cite{paul2024videolights} & SF+CLIP+BLIP & 52.94 & 61.96 & 41.05 & -- & -- \\
UniMD+Sync~\cite{zeng2024UniMD} & -- & -- & 63.98 & 44.46 & -- & -- \\
InternVideo2-1B~\cite{wang2024InternVideo2} & -- & --& 68.36 & 45.03 & -- & -- \\
EaTR~\cite{jang2023EaTR} & I3D & -- & 68.47 & 44.92 & -- & -- \\
Mr.BLIP~\cite{meinardus2024Mr.Blip} & BLIP-2 & 58.63 & 69.31 & 49.29 & -- & -- \\
InternVideo2-6B~\cite{wang2024InternVideo2} & -- & -- & 70.03 & 48.95 & -- & -- \\
FlashVTG~\cite{cao2024FlashVTG} & SF+CLIP & -- & 70.32 & \underline{49.87} & -- & -- \\
LLaVA-MR~\cite{lu2024llava} & CLIP+LLM & \underline{59.78} & \underline{70.65} & 49.58 & \textbf{69.96} & 39.66 \\
\midrule
\textbf{SMART} & CLIP+LLM & \textbf{61.09} & \textbf{72.26} & \textbf{52.17} & 69.47 & \textbf{40.43}\\

\bottomrule
\end{tabular}
}}
\vspace{-2mm}
\end{table}

\begin{figure*}[t]
\centerline{
  \includegraphics[width=\linewidth]{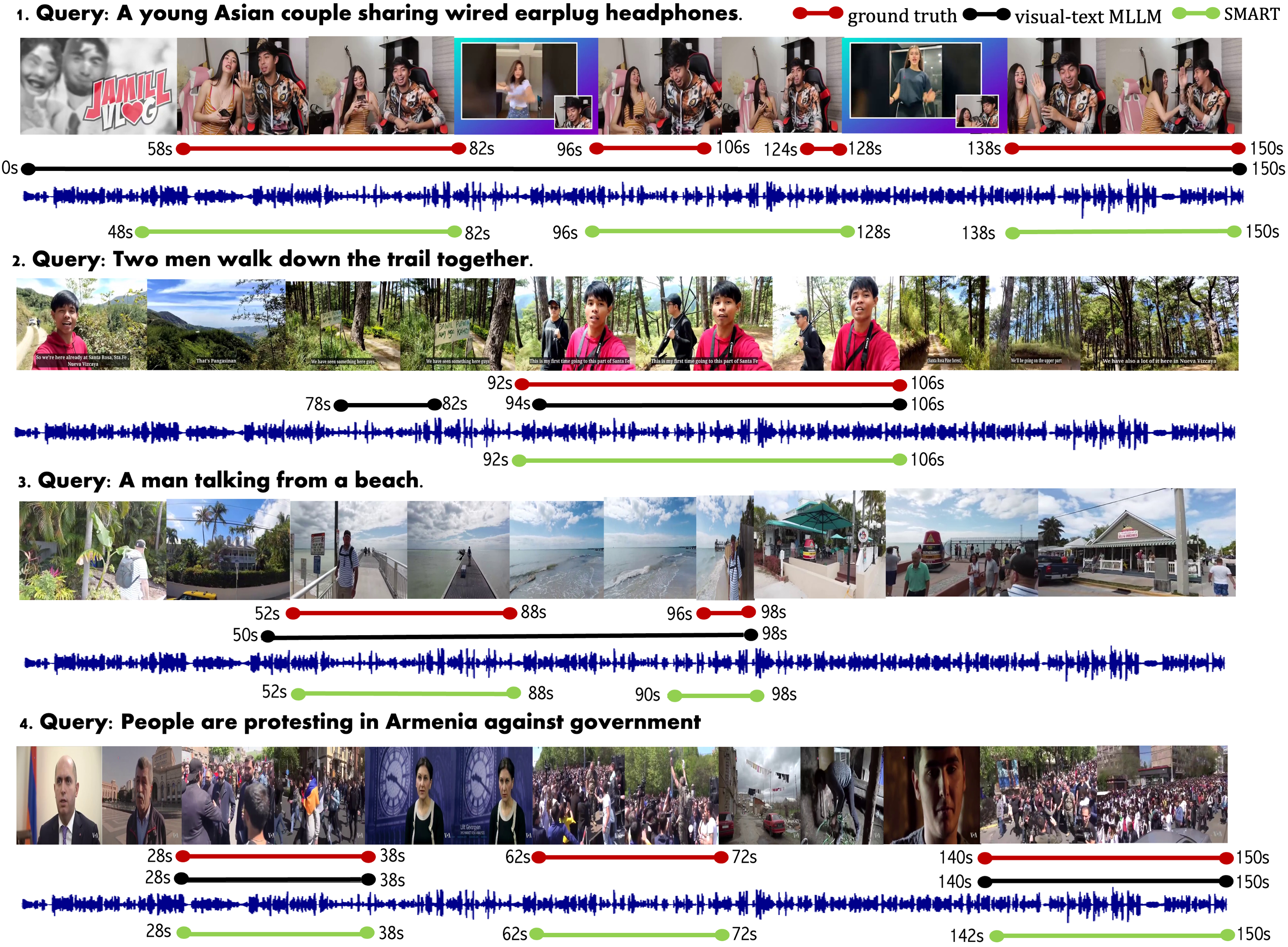} 
  \vspace{-2mm}
}
\caption{
{\bf Qualitative results on QVHighlights~\cite{lei2021detecting}:} 
{\normalfont This figure visualizes the predicted and ground-truth segments for query events. SMART outperforms the visual-text MLLM baseline by preserving shot-level consistency and leveraging audio cues for fine-grained temporal understanding. See text for details.}}
\label{fig:showcase}
\end{figure*}
\vspace{-5mm}

\subsection{Results}
\label{sec:results}

\noindent\textbf{Comparison to the State of the Art:}
We evaluate our approach on two widely used moment retrieval datasets: {\em Charades-STA} and {\em QVHighlights}. As shown in Table~\ref{tab:sota}, our model consistently surpasses prior methods across nearly all evaluation metrics on both benchmarks, demonstrating strong generalization regardless of dataset complexity or evaluation scenario.

Our proposed model, \textbf{SMART}, achieves substantial performance gains on both datasets, highlighting its robustness across videos ranging in duration from a few seconds to several minutes.

\noindent\textbf{QVHighlights:}
On the validation set, SMART surpasses previous approaches with improvements of \textbf{0.52\%} in R1@0.5 and \textbf{0.9\%} in R1@0.7. 
On the test set (Table~\ref{tab:sota}), SMART achieves the best results among the compared methods across the major reported metrics, achieving an R1@0.5 of \textbf{78.15}, R1@0.7 of \textbf{63.16}, mAP@0.5 of \textbf{70.76}, and mAP@0.75 of \textbf{55.54}. These results represent consistent and notable improvements over the strongest existing baselines, including LLaVA\textendash MR and Mr.BLIP. On the validation set, SMART also achieves the highest mIoU of \textbf{72.03}, outperforming all competing architectures built on BLIP and CLIP backbones. This demonstrates SMART’s superior ability to model fine-grained temporal structure and multimodal semantics.

\noindent\textbf{Charades\textendash STA:}
On the Charades\textendash STA test set, SMART also achieves clear performance gains, improving R1@0.5 by \textbf{1.61\%} and R1@0.7 by \textbf{2.59\%}. 
It further attains a new best mIoU of \textbf{61.09}, outperforming the previous leading method, Mr.BLIP (58.63), as well as other strong CLIP-based competitors such as CG\textendash DETR and VideoLights\textendash B\textendash pt. These consistent improvements across diverse benchmarks validate the effectiveness of our design choices, particularly the integration of LLM-driven semantic alignment with multimodal supervision. Overall, SMART achieves strong performance over prior methods across both short- and long-duration video benchmarks.

\begin{table*}[t]
\caption{Comprehensive ablation of the two main components in SMART on QVHighlights. Both audio integration and shot-aware token compression improve over the baseline, and their combination yields the best overall performance.
\vspace{-2mm}
}
\label{tab:ablation}
\centerline{
\begin{tabular}{ccc|cccc}
    \toprule
    &Audio&  Shot-aware Token Compression & R1@0.5 & R1@0.7 & mAP@0.5 & mAP@0.75 \\
    \midrule
    (a)&  & &76.52 & 63.23 & 68.99 & 55.25 \\
    (b)&\checkmark &   & 77.23 ({+0.71}) & 64.52 ({+1.29})  & 70.02 ({+1.03}) & 56.66 ({+1.41}) \\
    (c)& & \checkmark &  77.03 ({+0.51}) & 63.48 ({+0.25})  & 69.75 ({+0.76}) & 56.27 ({+1.02}) \\
    (d)& \checkmark& \checkmark & \textbf{78.65} ({+2.13}) & \textbf{65.03} ({+1.80})  & \textbf{70.46} ({+1.47}) & \textbf{56.72} ({+1.47}) \\
    \bottomrule
\end{tabular}%
}
\end{table*}

\medskip
\noindent\textbf{Qualitative Results:}
Fig.~\ref{fig:showcase} illustrates qualitative comparisons between ground-truth query segments and predicted intervals from baseline MLLM-based models (without audio) and our proposed \textbf{SMART}. In Example 1, competing MLLM models fail to maintain temporal precision due to the absence of shot-coherent token compression. In contrast, SMART leverages shot boundaries to retain the most relevant tokens, preserving contextual consistency within each shot and mitigating interference across shots. Moreover, by incorporating audio cues such as character dialogues, SMART captures fine-grained semantics more effectively, for example in the case of {\em ``sharing wired earplug headphones''}. These advantages enable SMART to deliver substantially more accurate predictions than competing approaches.

In Example 2, other MLLM-based models incorrectly predict the interval 78s–82s and miss the correct moment around 92s. This misalignment likely results from losing fine-grained visual cues—such as the presence of {\em ``two men''} during feature compression. In contrast, SMART applies shot-coherent compression that preserves essential information within each shot while filtering redundancy, enabling it to correctly identify the 92s–106s interval, which closely aligns with the ground truth.

In Example 3, for the query {\em ``A man talking from a beach''}, the keyword talking clearly requires audio understanding. Competing MLLM-based models output a continuous segment from 50s–98s. However, during 88s–96s the man is neither visible nor audible. By incorporating audio cues, SMART detects the absence of speech in that interval and instead predicts two separate segments of 52s–88s and 90s–98s. This matches the ground truth of 52s–88s and 96s–98s much better.

In Example 4, for the query {\em ``People are protesting in Armenia against the government''}, \rev{the ground-truth segment consists of multiple intervals:} 28s–38s, 62s–72s, and 140s–150s. While scene changes are clear, baseline MLLM models fail to detect the 62s–72s segment. In this portion, the visuals only depict a generic crowd without explicit references to {\em ``Armenia''}. However, the background audio includes a reporter explicitly mentioning {\em ``Armenia''} as the protest location. By leveraging this audio information, SMART captures the missing context and retrieves the correct moments more accurately. The Appendix provides additional qualitative examples with evaluation and detailed discussions.

To better understand where these gains come from, we next provide ablation, efficiency, and robustness analyses of the proposed audio integration and shot-aware compression design.


\subsection{Ablation Studies and Analysis}
\label{sec:ablation}
We further analyze SMART from four perspectives: the overall contribution of its main components, the role of audio under different query conditions, the stage-wise behavior of Shot-aware Token Compression (STC), and the resulting efficiency--effectiveness trade-off. We finally report hyperparameter studies on audio compression and STC configuration.

\textbf{Analysis of Audio Contribution.}
The benefit of audio varies across query types rather than being uniform across all queries. As shown in the supplementary query-type analysis, the majority of queries in QVHighlights are vision-sufficient, while a smaller subset are audio-helpful or strictly audio-dependent. Consistent with this distribution, SMART shows more noticeable improvements on audio-dependent and audio-helpful queries, whereas the gains on vision-sufficient queries are smaller but remain non-negative. This suggests that audio is not universally required, but provides complementary evidence when the target moment is at least partially grounded in acoustic cues.

We further compare several audio integration strategies in Table~\ref{tab:add_audio}. Among them, \emph{Overall Concatenation} performs best, achieving 77.23 R1@0.5, 64.52 R1@0.7, and 70.02 mAP@0.5. In contrast, \emph{Audio-Visual Fusion} performs worse, likely because the additional fusion parameters are harder to optimize under limited training data. \emph{Interleaved Integration} remains competitive in mAP@0.5 but is slightly weaker in recall, suggesting that enforcing fine-grained temporal correspondence may disrupt useful global audio context. Overall, these results indicate that a compact and globally coherent audio representation is more effective than strategies that impose fine-grained temporal alignment.

\textbf{Stage-wise Analysis of STC.}
To better understand the role of each stage in STC, we separately evaluate keyframe selection and intra-shot token filtering in Table~\ref{tab:stc_ab}. Keyframe selection alone provides a favorable trade-off: it improves coarse localization metrics such as R1@0.5 and mAP@0.75, although the gains are less consistent on stricter metrics such as R1@0.7 and mAP@0.5, indicating that temporal redundancy reduction alone is insufficient without subsequent semantic filtering.  In contrast, token filtering alone performs comparably on R1@0.7 but underperforms the audio-enhanced baseline on the other metrics, suggesting that aggressive token compression may discard useful contextual information along with redundant tokens. 
When the two stages are combined, SMART achieves the best overall performance, demonstrating that intra-shot token filtering is most effective when applied after shot-aware keyframe selection.

\begin{figure*}[t]
\centerline{
  \includegraphics[width=\linewidth]{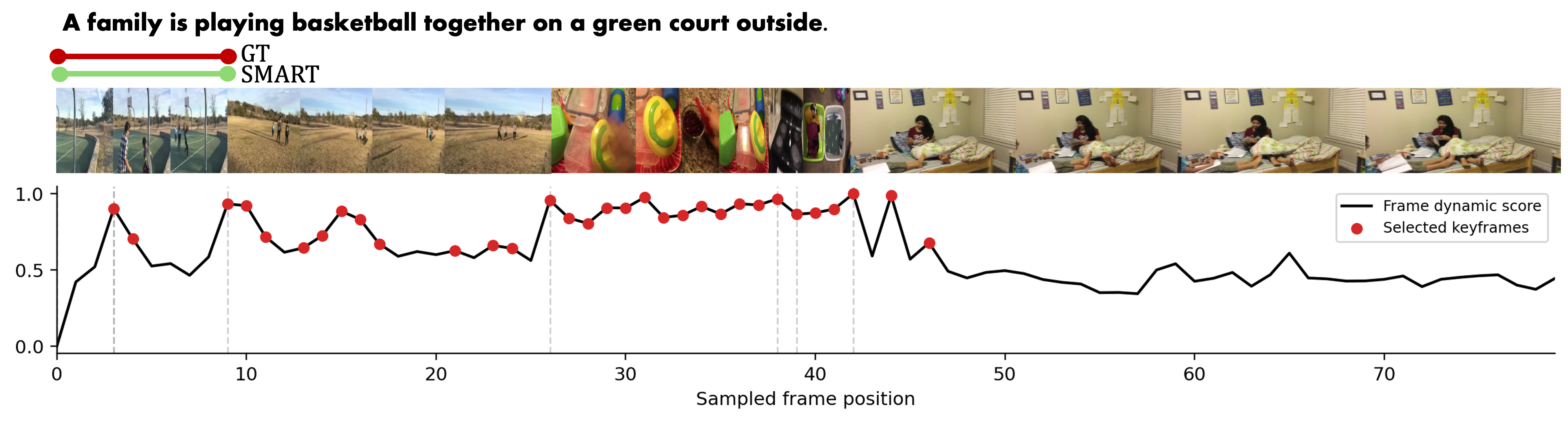} 
  \vspace{-4mm}
}
\caption{\textbf{Example of shot-aware keyframe selection on QVHighlights~\cite{lei2021detecting}.}
Our keyframe selection stage preserves representative keyframes from each shot as full frames, while the remaining non-keyframes are processed by the subsequent token filtering stage. This design retains the main temporal evidence while avoiding unnecessary redundancy.}\vspace{-2mm}
\label{fig:keyframe_select}
\end{figure*}

\begin{figure}
\centerline{
  \includegraphics[width=\linewidth]{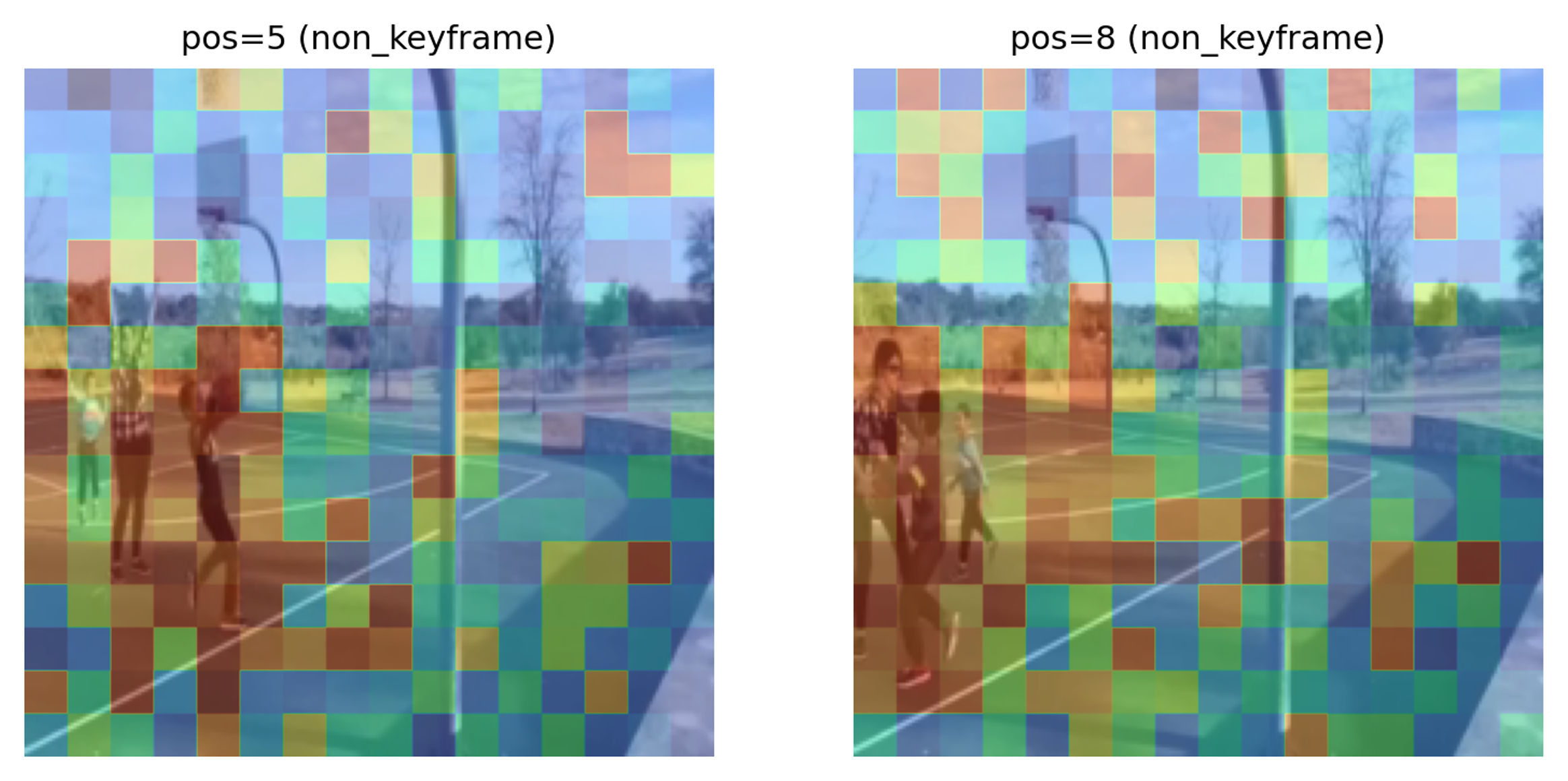} 
  \vspace{-4mm}
}

\caption{\textbf{
  Temporal feature variation heatmaps for two adjacent non-keyframes.}
  \rev{The heatmaps are computed from LayerNorm-normalized visual encoder patch features before STC compression.
  For each sampled frame $t$ and spatial patch $p$, we compute
  $H_t(p)=\|z_{t,p}-z_{t-1,p}\|_2$, where $z_{t,p}$ denotes the visual encoder feature of patch $p$ at frame $t$.
  The CLS token, if present, is excluded before reshaping the scores into a spatial grid.
  Warmer colors indicate stronger adjacent-frame feature changes.
  These maps are not attention maps; they visualize local temporal dynamics.
  The actual intra-shot token filtering operates on Q-Former visual query tokens after T5 projection, using
  $s(q)=\frac{1}{D}\sum_{d=1}^{D}\operatorname{Std}_{t\in\mathcal{S}}(h_{t,q,d})$ within each shot $\mathcal{S}$, and retains tokens with the highest
  $s(q)$ values.
  }}

\vspace{-2mm}

\label{fig:heatmap}
\end{figure}

Figure~\ref{fig:keyframe_select} provides a qualitative example of this process. Representative keyframes are preserved as full frames, while non-keyframes are passed to the subsequent token filtering stage. Figure~\ref{fig:heatmap} further visualizes the temporal feature variation within non-keyframes: tokens with stronger temporal dynamics are retained, whereas weakly varying regions are more likely to be suppressed. Together, these results support STC as a two-stage compression mechanism rather than a uniform reduction strategy.

\textbf{Efficiency Analysis.}
We next examine the computational effect of different STC variants in Table~\ref{tab:efficiency}. As expected, \emph{Token filtering only} yields the largest raw efficiency gains, reducing the token retention ratio to 56.00\%, peak GPU memory to 19.42~GB, and end-to-end latency to 2168.47~ms. However, this aggressive compression also leads to weaker retrieval performance than the full model. SMART is slightly more expensive because selected keyframes are preserved as full frames and only non-keyframes are compressed. Even so, SMART still reduces peak GPU memory from 21.94~GB to 20.72~GB and end-to-end latency from 2451.49~ms to 2224.04~ms relative to the audio-enhanced baseline, while maintaining the strongest retrieval performance. This suggests that SMART provides a more favorable efficiency--effectiveness trade-off than applying token filtering alone.

\begin{figure}[t]
\centering

\begin{minipage}{0.48\linewidth}
    \centering
    \includegraphics[width=\linewidth]{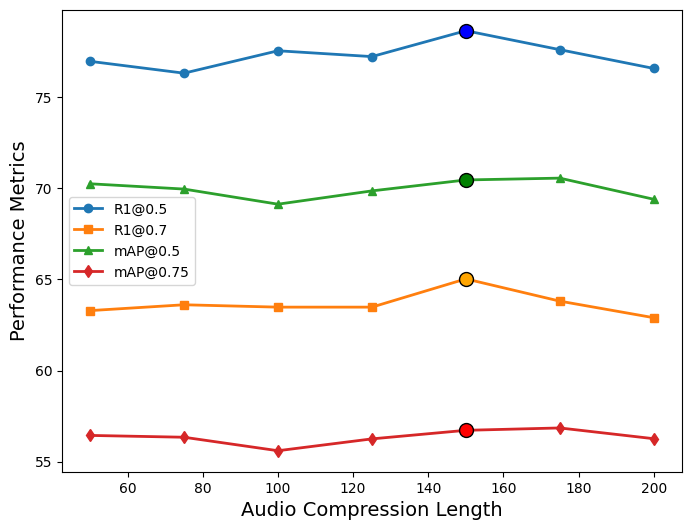}
    \vspace{-3mm}
    \caption{Impact of {\bf audio compression length $L$}
    of the Overall Concatenation strategy on QVHighlights, where the circular marker at $L=150$ indicates the optimal setting for all metrics.}
    \label{fig:audio_len}
\end{minipage}
\hfill
\begin{minipage}{0.48\linewidth}
    \centering
    \includegraphics[width=\linewidth]{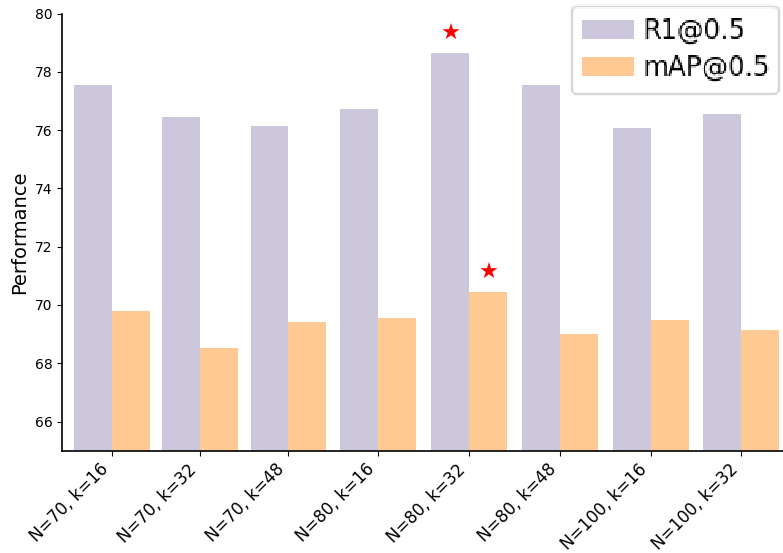}
    \vspace{-3mm}
    \caption{Determining the optimal {\bf frame number $N$}
    and {\bf top-$k$ keyframe} for Shot-aware Token Compression (STC) on QVHighlights, where red stars mark the best configuration ($N =80$, $k=32$).}
    \label{fig:N_k}
\end{minipage}

\end{figure}

\textbf{Hyperparameter Studies.}
We finally study two implementation choices on QVHighlights. First, Figure~\ref{fig:audio_len} analyzes the audio compression length $L$ under the \emph{Overall Concatenation} strategy. We observe that $L=150$ gives the best overall balance across recall and mAP metrics, suggesting that overly short audio sequences lose useful context, whereas overly long ones introduce redundancy. Second, Figure~\ref{fig:N_k} studies the total frame number $N$ and the number of selected keyframes $k$ for STC. The best configuration is obtained at $N=80$ and $k=32$, which balances visual coverage and sequence efficiency. Too few keyframes lead to excessive compression, whereas too many reintroduce redundancy without consistent gains.

\begin{table*}[t]
\caption{\textbf{Stage-wise ablation of STC on QVHighlights.}
Keyframe selection consistently helps, while token filtering alone is less stable.
Their combination in SMART achieves the best overall performance.} \vspace{-2mm}
\label{tab:stc_ab}
\centering
\begin{tabular}{lcccccc}
\toprule
Variant & Keyframe Selection & Intra-shot Token Filtering & R1@0.5 & R1@0.7 & mAP@0.5 & mAP@0.75 \\
\midrule
Keyframe only & \checkmark &  & 77.35 & 63.81 & 69.95 & 56.76 \\
Token filtering only &  & \checkmark & 76.90 & 64.19 & 69.21 & 55.94 \\
SMART & \checkmark & \checkmark & 78.65 & 65.03 & 70.46 & 56.72 \\
\bottomrule
\end{tabular}
\vspace{-2mm}
\end{table*}

\begin{table*}[t]
\centering

\begin{minipage}{0.49\textwidth}
\centering
\caption{\textbf{Efficiency comparison on QVHighlights.}\\
Token filtering only achieves the largest efficiency gains, while SMART offers a stronger efficiency--effectiveness trade-off.}
\label{tab:efficiency}
\vspace{-2mm}

\resizebox{\linewidth}{!}{
\begin{tabular}{lcccc}
\toprule
\textbf{Method} & 
\makecell{\textbf{Token Kept} \\ \textbf{\%}} &
\makecell{\textbf{Peak GPU Mem} \\ \textbf{GB}} &
\makecell{\textbf{E2E Latency} \\ \textbf{ms}} &
\makecell{\textbf{Throughput} \\ \textbf{samples/s}} \\
\midrule
Baseline with audio & 100   & 21.94 & 2451.49 & 0.4331 \\
Keyframe only       & 88.75 & 21.29 & 2368.54 & 0.4456 \\
Token filtering only& 56.00 & 19.42 & 2168.47 & 0.4924 \\
SMART               & 78.89 & 20.72 & 2224.04 & 0.4750 \\
\bottomrule
\end{tabular}
}
\end{minipage}
\hfill
\begin{minipage}{0.49\textwidth}
\centering
\caption{\textbf{Ablation study of different audio concatenation strategies}\\
on the QVHighlights dataset.}
\label{tab:add_audio}
\vspace{-2mm}

\resizebox{\linewidth}{!}{
\begin{tabular}{lcccc}
\toprule
Audio & R1@0.5 & R1@0.7 & mAP@0.5 & mAP@0.75 \\
\midrule
Overall Concatenation & \textbf{77.23} & \textbf{64.52} & \textbf{70.02} & 56.66 \\
Audio-Visual Fusion & 24.84 & 16.23 & 20.47 & 12.26 \\
Interleaved Alignment & 76.71 & 63.16 & 70.01 & 56.12 \\
Voice + Ambient Sound & 76.00 & 63.61 & 69.82 & 56.83 \\
\bottomrule
\end{tabular}
}
\end{minipage}

\vspace{-2mm}
\end{table*}

\section{Conclusion}

We introduce \textbf{Shot-aware Multimodal Audio-enhanced Retrieval of Temporal Segments (SMART)}, a multimodal moment retrieval framework that improves temporal localization by incorporating complementary audio cues together with a \textbf{Shot-aware Token Compression} strategy. SMART first performs shot-aware keyframe selection to preserve representative frames as full visual observations, and then applies token filtering only to non-keyframes to reduce redundant visual content while retaining temporally informative regions. This two-stage design balances efficiency and effectiveness, enabling more efficient long-sequence processing while preserving retrieval performance. In parallel, compact audio representations provide complementary semantic cues that are particularly helpful when the target moment is at least partially grounded in acoustic evidence. Extensive experiments on Charades-STA and QVHighlights show that SMART consistently improves over strong baselines while achieving favorable performance--efficiency trade-offs.

\medskip
\noindent\textbf{Limitations:} Despite its strong performance, SMART still has several limitations. First, the current audio branch relies on compressed global audio context rather than explicitly modeling fine-grained audio--visual temporal alignment, which may limit its ability to capture subtle audio-event timing in complex scenes. Second, although the proposed shot-aware compression preserves selected keyframes and only compresses non-keyframes, subtle visual details may still be weakened under aggressive compression or highly dynamic motion patterns. Third, while our additional analyses show that SMART is reasonably robust to moderate shot-boundary perturbations and can transfer its visual compression design to another backbone, its behavior under larger distribution shifts or broader open-world settings remains an important direction for future work.

\medskip
\noindent\textbf{Future Work:} SMART offers several opportunities for improvement. Dense audio-text alignment could enhance grounding in complex scenes, while extending SMART to open-world or zero-shot retrieval could improve generalization to unseen events and domains. Integrating additional modalities or hierarchical temporal modeling may further strengthen robust, scalable multimodal moment retrieval.

\noindent
{\bf Acknowledgment:} Felix X.-F. Ye is grateful for partial support from seed funding by the
Center for Emerging Artificial Intelligence Systems at the
University at Albany. This research was supported in part by the AI Computing Cluster at the University at Albany.


\clearpage
\appendices
\section*{Supplementary Material}

\section{Additional Ablation on Audio Design}

\noindent\textbf{Choice of audio encoder.}
BEATs is selected as the preferred audio encoder due to its strong and consistent performance on QVHighlights. As shown in Table~\ref{tab:encoder}, it achieves the best results in three of the four metrics, including R1@0.7, mAP@0.5, and mAP@0.75. Although Encodec slightly outperforms BEATs in R1@0.5, the margin is small and does not translate into stronger overall performance. These results suggest that BEATs provides a more balanced trade-off between recall and precision for multimodal moment retrieval. All encoder comparisons are conducted without additional components such as shot-aware token compression.

\medskip
\noindent\textbf{Audio shift analysis.}
Table~\ref{tab:audio-shift} evaluates the sensitivity of SMART to temporal perturbations in the audio stream. The aligned model achieves 78.65 R1@0.5 and 70.46 mAP@0.5. Under audio shifts of $\pm$2s and $\pm$5s, the performance decreases slightly to the range of 77.74--77.94 in R1@0.5 and 69.86--69.99 in mAP@0.5. Randomly shuffling the audio chunks yields 77.87 R1@0.5 and 69.99 mAP@0.5. Overall, the degradation remains within about 0.7--0.9 points in R1@0.5 and 0.5--0.6 points in mAP@0.5, indicating that SMART is reasonably robust to coarse temporal perturbations in the audio stream. These results suggest that audio primarily provides complementary contextual information, rather than serving as a precise temporal alignment signal.

\section{Additional Analysis of STC}

\noindent\textbf{Comparison of visual compression strategies.}
Table~\ref{tab:token_compression} compares different visual compression strategies on Charades-STA. Without compression, the model achieves 68.52 R1@0.5, 47.29 R1@0.7, and 58.16 mIoU. Applying average pooling slightly improves the performance to 69.43 R1@0.5, 48.17 R1@0.7, and 58.77 mIoU, suggesting that moderate token reduction can already help by shortening the visual sequence. Our proposed shot-aware compression performs best across all metrics, reaching 70.39 R1@0.5, 49.03 R1@0.7, and 59.47 mIoU. These results suggest that the gain does not come from compression alone, but from preserving representative keyframes as full frames while compressing only non-keyframes, which reduces redundancy while retaining informative temporal evidence.

\noindent\textbf{Robustness to shot-boundary perturbations.}
Table~\ref{tab:shot_boundary} studies the effect of different shot-boundary settings on QVHighlights. Using the default TransNetV2 boundaries yields the best performance, with 78.65 R1@0.5 and 70.46 mAP@0.5. Replacing them with fixed-length pseudo-shots leads to a more noticeable drop, reducing the performance to 77.74 R1@0.5 and 69.95 mAP@0.5. In contrast, a mild boundary perturbation of $\pm 5$ frames causes only limited degradation, with 78.00 R1@0.5 and 69.96 mAP@0.5, while boundary corruption by merging or dropping 20\% of the detected boundaries still maintains 77.81 R1@0.5 and 70.03 mAP@0.5. These results suggest that SMART benefits from reasonable shot boundaries, but does not rely on perfect shot segmentation.

\section{Query-Type Analysis}
\begin{figure}[t] %
\centerline{
  \includegraphics[width=0.8\linewidth]{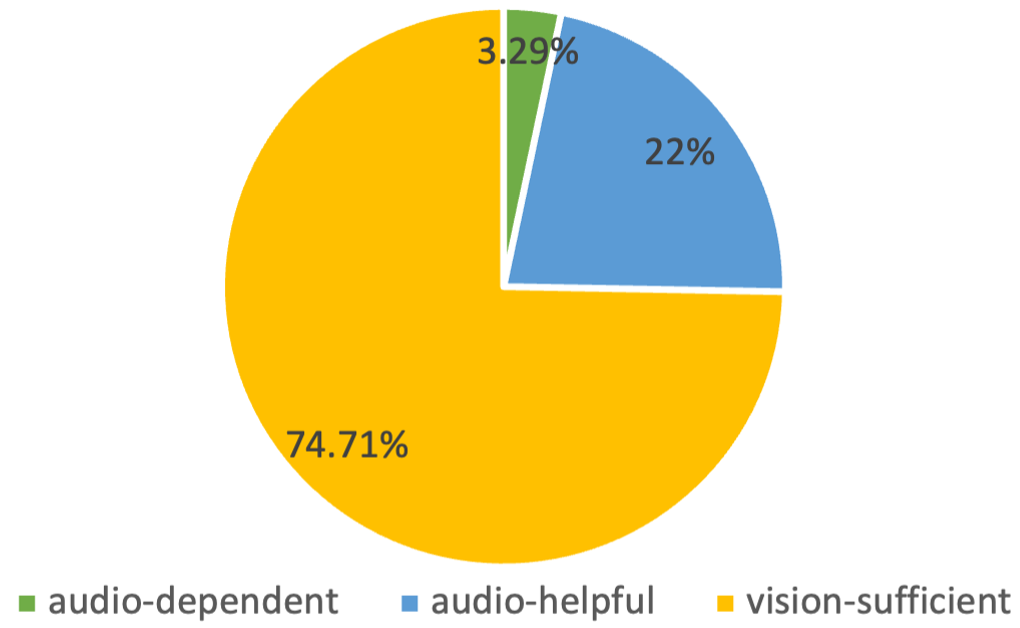} 
  \vspace{-2mm}
}
\caption{\textbf{Distribution of query types on QVHighlights.}
Most queries are vision-sufficient, while a smaller subset are audio-helpful or audio-dependent.} \vspace{-2mm}
\label{fig:query_dist}

\end{figure}

Figure~\ref{fig:query_dist} and Table~\ref{tab:query_perform} provide a query-type analysis on QVHighlights. As shown in Figure~\ref{fig:query_dist}, the majority of queries are \emph{vision-sufficient} (74.71\%), whereas \emph{audio-helpful} and \emph{audio-dependent} queries account for 22.00\% and 3.29\%, respectively. Table~\ref{tab:query_perform} further shows that SMART brings substantially larger gains on audio-relevant queries. On \emph{audio-dependent} queries, SMART improves R1@0.5 by 8.83 points (58.82 $\rightarrow$ 67.65) and mAP@0.5 by 6.38 points (61.76 $\rightarrow$ 68.14). On \emph{audio-helpful} queries, the gains are 4.72 points in R1@0.5 (68.55 $\rightarrow$ 73.27) and 3.06 points in mAP@0.5 (62.30 $\rightarrow$ 65.36). By comparison, the gains on \emph{vision-sufficient} queries are smaller, at 0.45 points in R1@0.5 (79.73 $\rightarrow$ 80.18) and 1.47 points in mAP@0.5 (70.17 $\rightarrow$ 71.64). These results suggest that audio is not uniformly critical for all queries, but it provides clear benefits when the query is at least partially dependent on audio cues. 

\section{Additional Qualitative Results}

\begin{table}[t]
  \centering
  \caption{\textbf{Ablation study of audio encoder choice} on the QVHighlights dataset.}
  \label{tab:encoder}
  \begin{tabular}{lcccc}
    \toprule
    \textbf{Encoder} & \textbf{R1@0.5} & \textbf{R1@0.7} & \textbf{mAP@0.5} & \textbf{mAP@0.75} \\
    \midrule
    BEATs~\cite{chen2022BEATs}       & 77.23 & \textbf{64.52} & \textbf{70.02} & \textbf{56.66} \\
    Whisper~\cite{radford2022robust} & 77.16 & 62.71          & 68.89          & 55.33 \\
    Encodec~\cite{defossez2022high}  & \textbf{77.68} & 63.10 & 68.19          & 54.51 \\
    \bottomrule
  \end{tabular}
\end{table}

\begin{table}[t]
\caption{\textbf{Audio shift analysis on QVHighlights.}
All perturbed audio settings lead to slight performance drops relative to the aligned SMART model, indicating that SMART benefits from complementary audio information while remaining robust to coarse temporal perturbations.} \vspace{-2mm}
\label{tab:audio-shift}
\centering
\begin{tabular}{lcc}
\toprule
\textbf{Audio Setting} & \textbf{R1@0.5} & \textbf{mAP@0.5} \\
\midrule
Aligned (0s) & 78.65 & 70.46 \\
Shift +2s    & 77.81 & 69.93 \\
Shift -2s    & 77.74 & 69.86 \\
Shift +5s    & 77.94 & 69.98 \\
Shift -5s    & 77.81 & 69.93 \\
Shuffle      & 77.87 & 69.99 \\
\bottomrule
\end{tabular}
\end{table}

\begin{table}[t]
  \centering
  \caption{\textbf{Ablation study} of different visual token compression methods on the Charades-STA dataset.
  \vspace{-2mm}
  }
  \label{tab:token_compression}
  \resizebox{\linewidth}{!}{
    \begin{tabular}{lccc}
      \toprule
      \textbf{Compression Method} & \textbf{R1@0.5} & \textbf{R1@0.7} & \textbf{mIoU} \\
      \midrule
      No Compression   & 68.52 & 47.29 & 58.16 \\
      Average Pooling  & 69.43 & 48.17 & 58.77 \\
      Ours             & \textbf{70.39} & \textbf{49.03} & \textbf{59.47} \\
      \bottomrule
    \end{tabular}
  }
\end{table}

\begin{figure*}[t] 
\centerline{
  \includegraphics[width=\linewidth]{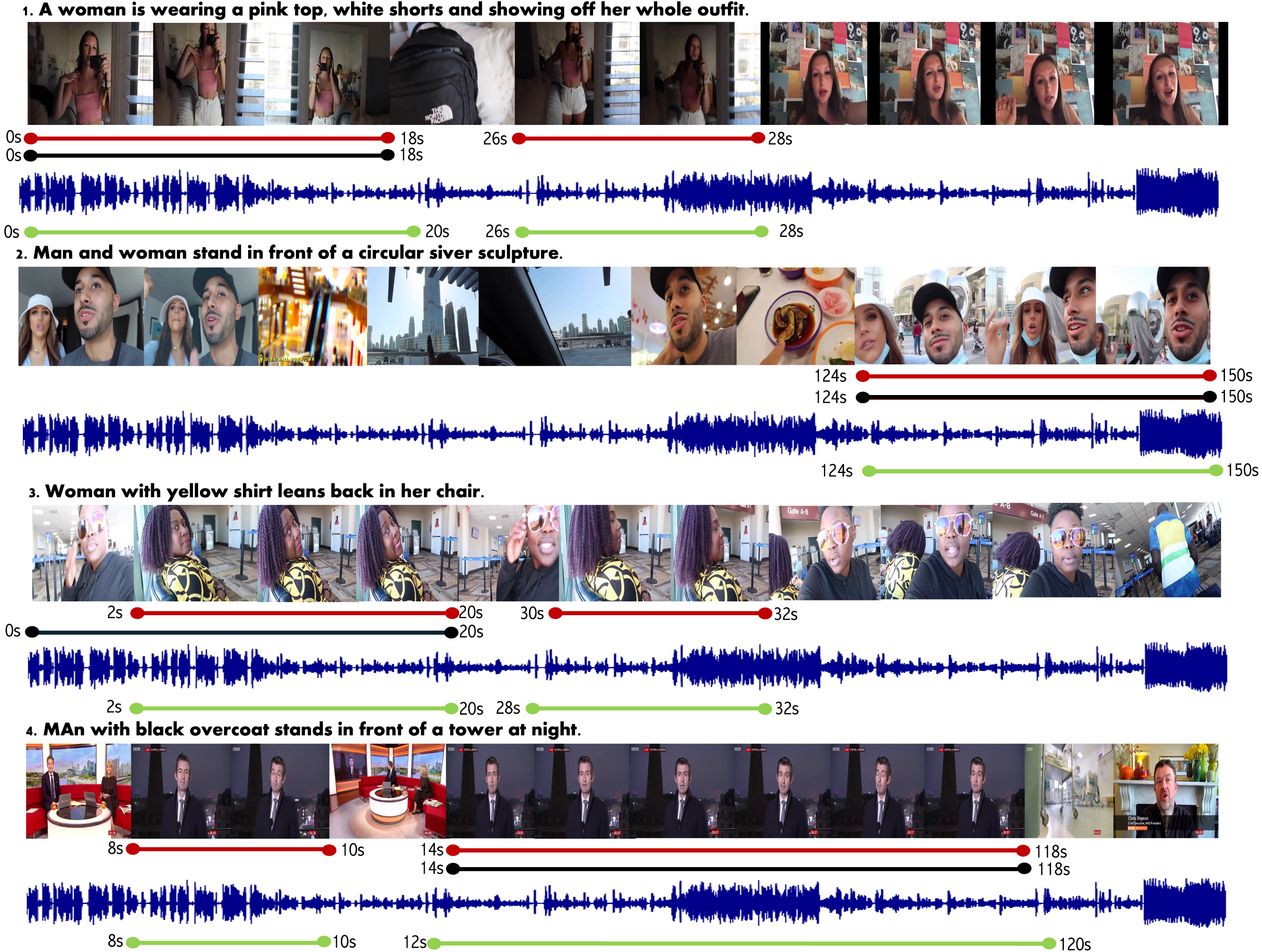} 
  \vspace{-2mm}
}
\caption{
{\bf More Qualitative results} on  QVHighlights~\cite{lei2021detecting}: Predicted and ground truth segments are shown for comparison.}
\label{fig:showcase2}

\end{figure*}

\begin{figure*}[t] %
\centerline{
  \includegraphics[width=\linewidth]{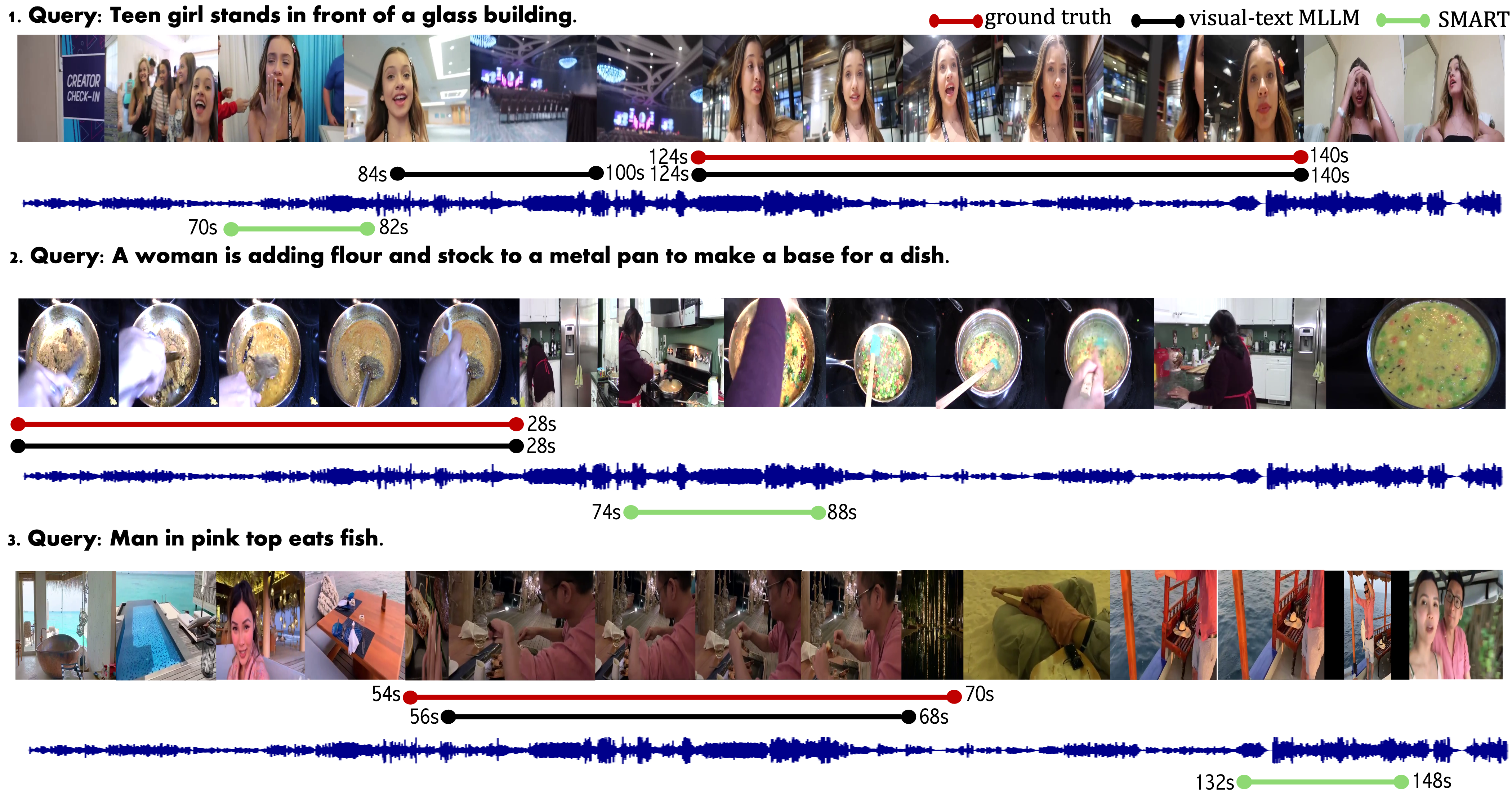} 
  \vspace{-2mm}
}
\caption{
{\bf \rev{Failure cases} on  QVHighlights~\cite{lei2021detecting}: Predicted and ground truth segments are shown for comparison.}}
\label{fig:bad_exp}

\end{figure*}

We provide \rev{additional} qualitative results shown in Figure~\ref{fig:showcase2}.

\medskip
\noindent
\textbf{Example 1}: Query: ``A woman is wearing a pink top, white shorts and showing off her whole outfit.''

The ground truth consists of two disjoint segments (0s–18s and 26s–28s). 
The visual-text MLLM captures the first segment but misses the short second segment near the end, likely due to its reliance on visual cues that emphasize prominent poses while overlooking subtle continuation signals. 
In contrast, SMART recovers both segments: it slightly extends the first segment to around 20s and successfully identifies the short second segment. 
This improvement is partly due to the integration of audio cues (e.g., narration or ambient signals) that align with the full semantic event, as well as shot-aware token compression that preserves informative visual tokens across dynamic frames.

\medskip
\noindent
\textbf{Example 2}: Query: ``Man and woman stand in front of a circular silver sculpture.''

This example corresponds to a long and visually static segment (124s–150s), where both the visual-text MLLM and SMART successfully capture the temporal extent of the event. 
Despite the limited visual variation, the scene remains semantically consistent over time, making it relatively easy for both models to localize. 
Notably, SMART maintains stable predictions in such low-motion scenarios, benefiting from complementary audio cues (e.g., consistent ambient sounds or background speech) and shot-aware token compression, which helps preserve semantically relevant but visually subtle frames.

\medskip
\noindent
\textbf{Example 3}: Query: ``Woman with yellow shirt leans back in her chair.''

The ground truth consists of two separated segments (2s–20s and 30s–32s). 
The visual-text MLLM captures the first segment but misses the short second segment near the end, likely because the initial leaning motion is visually salient while the later frames appear less distinctive. 
In contrast, SMART successfully recovers both segments, including the short second interval. 
This improvement can be attributed to complementary audio cues (e.g., chair movement or subtle acoustic signals) that indicate the action is still ongoing, together with shot-aware token compression that preserves informative motion-related tokens while filtering redundant background content.

\medskip
\noindent
\textbf{Example 4}: Query: ``Man with black overcoat stands in front of a tower at night.''

The ground truth consists of two segments (8s–10s and 14s–118s). 
The visual-text MLLM fails to capture the full temporal extent, producing truncated predictions and missing part of the long second segment. 
In contrast, SMART better captures the long second segment while also identifying the short initial segment. 
This improvement is supported by complementary audio cues (e.g., consistent ambient sound or background speech) that reinforce scene continuity, as well as shot-aware token compression that preserves important yet low-variance visual tokens throughout the shot.

We also provide representative failure cases in Figure~\ref{fig:bad_exp} to clarify the boundary conditions of SMART. These failure cases mainly occur when the query is primarily visually grounded and the additional audio stream provides weakly relevant or misleading context. In such cases, the audio branch may interfere with temporal grounding rather than improve it.

\medskip
\noindent
\textbf{Example 1}: Query: ``Teen girl stands in front of a glass building.''

This query is primarily visual and contains little explicit audio-related information. The ground-truth moment spans 124s--140s, while the visual-text MLLM predicts an earlier partially related segment from 84s to 100s. SMART is further distracted and predicts an even earlier short segment from 70s to 82s. A likely reason is that the added audio introduces irrelevant or weakly related acoustic context, which does not help identify the visually defined target moment and instead biases the model toward an incorrect earlier segment.

\medskip
\noindent
\textbf{Example 2}: Query: ``A woman is adding flour and stock to a metal pan to make a base for a dish.''

This query also depends mainly on fine-grained visual details rather than audio cues. The ground-truth segment occurs at the beginning of the video and ends at 28s, which is correctly captured by the visual-text MLLM. However, SMART predicts a later segment from 74s to 88s. Since multiple cooking stages share similar kitchen sounds, the added audio may provide ambiguous or misleading context instead of helping distinguish the correct step. As a result, the model is attracted to a later but semantically related cooking segment.

\medskip
\noindent
\textbf{Example 3}: Query: ``Man in pink top eats fish.''

This query is largely vision-sufficient and does not explicitly require audio for localization. The ground-truth interval is 54s--70s, and the visual-text MLLM predicts a nearby but incomplete segment from 56s to 68s. In contrast, SMART predicts a much later segment from 132s to 148s. This suggests that the added audio does not provide useful disambiguation and may instead interfere with temporal grounding, especially when similar visual semantics reappear later in the video and the soundtrack is not strongly tied to the queried action.

\begin{table}[t]
\caption{\textbf{Performance breakdown by query type on QVHighlights.}
SMART brings the largest gains on audio-dependent and audio-helpful queries, while still maintaining competitive performance on vision-sufficient queries.} \vspace{-2mm}
\label{tab:query_perform}
\centering
\begin{tabular}{lcccc}
\toprule
& \multicolumn{2}{c}{R1@0.5} & \multicolumn{2}{c}{mAP@0.5} \\
\cmidrule(lr){2-3} \cmidrule(lr){4-5}
\textbf{Query Type} & Baseline & SMART & Baseline & SMART \\
\midrule
audio-dependent   & 58.82 & 67.65 & 61.76 & 68.14 \\
audio-helpful     & 68.55 & 73.27 & 62.30 & 65.36 \\
vision-sufficient & 79.73 & 80.18 & 70.17 & 71.64 \\
\bottomrule
\end{tabular}
\end{table}


\begin{table}[t]
\caption{\textbf{Robustness to shot-boundary perturbations on QVHighlights.}
The default TransNetV2 boundaries give the best results, while pseudo-shots cause a larger drop. Mild perturbation or partial boundary removal leads to only limited degradation, suggesting that SMART benefits from reasonable shot boundaries without relying on perfect shot segmentation.} \vspace{-2mm}
\label{tab:shot_boundary}
\centering
\begin{tabular}{lcc}
\toprule
\textbf{Boundary Setting} & R1@0.5 & mAP@0.5 \\
\midrule
TransNetV2 boundary                     & 78.65 & 70.46 \\
fixed-length pseudo-shots              & 77.74 & 69.95 \\
boundary perturbation ($\pm$5 frames)  & 78.00 & 69.96 \\
boundary corruption (merge/drop 20\%)  & 77.81 & 70.03 \\
\bottomrule
\end{tabular}
\end{table}

\section{\rev{Ablation on Audio Feature Pooling}}

We conduct an ablation study on QVHighlights to compare different pooling strategies for compressing BEATs audio features, including average pooling, max pooling, and attention-weighted pooling. As shown in Table~\ref{tab:audio_pooling}, average pooling achieves the best overall performance, obtaining $78.65$ R1@0.5, $65.03$ R1@0.7, and $70.46$ mAP@0.5. In contrast, max pooling and attention-weighted pooling lead to consistently lower retrieval accuracy on the primary evaluation metrics.

Although attention-weighted pooling slightly improves mAP@0.75, its performance on R1@0.5, R1@0.7, and mAP@0.5 remains inferior to average pooling. Max pooling performs the worst overall, likely because it tends to overemphasize transient high-energy background sounds that are weakly correlated with the queried moments.

Based on these results, we adopt average pooling as our default audio compression strategy, as it provides the best overall trade-off between effectiveness, stability, and computational simplicity.

\begin{table}[t]
  \centering
  \caption{\rev{Ablation of audio feature pooling strategies on QVHighlights.}}
  \label{tab:audio_pooling}
  \begin{tabular}{lcccc}
  \toprule
  Pooling Strategy & R1@0.5 & R1@0.7 & mAP@0.5 & mAP@0.75 \\
  \midrule
  Max pooling & 77.10 & 63.94 & 69.53 & 56.55 \\
  Average pooling & \textbf{78.65} & \textbf{65.03} & \textbf{70.46} & 56.72 \\
  Attention pooling & 76.65 & 63.55 & 69.49 & \textbf{56.88} \\
  \bottomrule
  \end{tabular}
  \end{table}

\section{Cross-Backbone Validation}

To further examine whether the gains depend on a specific backbone, we evaluate our visual compression design on an alternative Video-LLaMA backbone. Since Video-LLaMA~\cite{zhang2023videollama} already incorporates audio in its native multimodal architecture, we do not introduce an additional audio branch in this auxiliary experiment. Instead, we focus on transferring the visual compression components of our method, namely keyframe selection and non-keyframe token filtering. As shown in Table~\ref{tab:extra_backbone}, VideoLLaMA$^{*}$ refers to the original Video-LLaMA augmented with keyframe selection and token filtering. This modification consistently improves performance on Charades-STA, raising R1@0.5 from 48.93 to 53.26, R1@0.7 from 27.64 to 30.12, and mIoU from 31.27 to 32.94. These gains suggest that the benefit of the proposed visual compression design is not limited to the backbone used in the main paper. At the same time, since Video-LLaMA does not follow the same shot-aware pipeline as SMART, we view this experiment as a cross-backbone validation of the visual compression mechanism rather than a fully identical reimplementation of the entire SMART framework. 

\begin{table}[ht]
\caption{\textbf{Cross-backbone validation on Charades-STA.}
Adding keyframe selection and token filtering to Video-LLaMA consistently improves performance, suggesting that the proposed visual compression design is not tied to the backbone used in the main paper. \vspace{-2mm}}
\label{tab:extra_backbone}
\centering
\begin{tabular}{lccc}
\toprule
\textbf{Method} & R1@0.5 & R1@0.7 & mIoU \\
\midrule
VideoLLaMA                     & 48.93 & 27.64 & 31.27 \\
VideoLLaMA$^{*}$              & 53.26 & 30.12 & 32.94 \\
\bottomrule
\end{tabular}
\end{table}


\end{document}